\documentclass[letterpaper]{article} % DO NOT CHANGE THIS
\usepackage{aaai2026}  % DO NOT CHANGE THIS
\usepackage{times}  % DO NOT CHANGE THIS
\usepackage{helvet}  % DO NOT CHANGE THIS
\usepackage{courier}  % DO NOT CHANGE THIS
\usepackage[hyphens]{url}  % DO NOT CHANGE THIS
\usepackage{graphicx} % DO NOT CHANGE THIS
\usepackage{natbib}  % DO NOT CHANGE THIS AND DO NOT ADD ANY OPTIONS TO IT
\usepackage{caption} % DO NOT CHANGE THIS AND DO NOT ADD ANY OPTIONS TO IT
\usepackage{pifont}
\newcommand{\cmark}{\ding{51}}
\newcommand{\xmark}{\ding{55}}
\usepackage{amssymb, amsmath}
\usepackage{placeins}
\usepackage{booktabs} % 用于三线表
\usepackage{xcolor}   % 用于颜色高亮
\usepackage{multirow} % 用于多行合并
\usepackage{ifpdf} % for conditional pdfinfo
\usepackage{algorithm}
\usepackage{algorithmic}

\usepackage{newfloat}
\usepackage{listings}
\DeclareCaptionStyle{ruled}{labelfont=normalfont,labelsep=colon,strut=off} % DO NOT CHANGE THIS
\floatstyle{ruled}
\newfloat{listing}{tb}{lst}{}
\floatname{listing}{Listing}
\nocopyright 
\title{
MemOVCD: Cross-Temporal Memory-Conditioned Inference for Training-Free Open-Vocabulary Change Detection
}
\author{
    Zuzheng Kuang\equalcontrib,
    Honghao Chang\equalcontrib,
    Boqiang Liang,
    Haoqian Wang,
    Lijun He,
    Fan Li,
    Haixia Bi\thanks{Corresponding author.}
}
\affiliations{
    School of Information and Communications Engineering, Xi’an Jiaotong University\\
    Xi’an 710049, China\\
    kzz794466014@stu.xjtu.edu.cn,
    2193712652@stu.xjtu.edu.cn,
    lbq15536293860@stu.xjtu.edu.cn, \\
    1226538518@stu.xjtu.edu.cn,
    lijunhe@mail.xjtu.edu.cn,
    lifan@mail.xjtu.edu.cn,
    haixia.bi@xjtu.edu.cn
}

\usepackage{bibentry}
\begin{document}

\maketitle

\begin{abstract}

Open-vocabulary change detection (OVCD) seeks to identify semantic changes in bi-temporal remote sensing imagery with text-specified target categories. Representative foundation-model pipelines decode the two timestamps independently before comparing their predictions, leaving cross-temporal evidence unavailable to semantic decoding. Moreover, current patch-wise inference can undermine global semantic continuity, yielding fragmented predictions. To address these limitations, we put forward MemOVCD, a training-free framework which reformulates OVCD as cross-temporal memory-conditioned inference. Specifically, MemOVCD casts bi-temporal analysis as a bidirectional two-frame tracking problem and exploits the memory mechanism of video SAM to propagate semantic candidate regions across timestamps. Their features are pooled and averaged into a local exemplar token. Subsequently, a shared memory, whose number of prototypes is determined by local-exemplar availability, aggregates regional features from the unlabeled target collection. The local and shared tokens are then jointly used to condition the subsequent semantic segmentation. We further devise a global-local adaptive rectification strategy, which adaptively fuses local and global predictions to compensate for incomplete spatial support when processing large images while preserving fine-grained details. MemOVCD and its inductive variant L-MemOVCD show favorable performance on five benchmarks, and ablations and analyses further validate its effectiveness under evaluated settings. Our code will be publicly available.

\end{abstract}

\section{Introduction}

Open-vocabulary change detection (OVCD) aims to localize and categorize semantic changes specified by text queries, without retraining category-specific models \citep{zhuang2025ov}. Compared with conventional closed-set change detection, OVCD provides substantially greater flexibility for remote sensing applications, where categories of interest vary across regions and practical scenarios \citep{li2026dynamicearth}.

\begin{figure}[!t]\centering
\includegraphics[width=\linewidth]{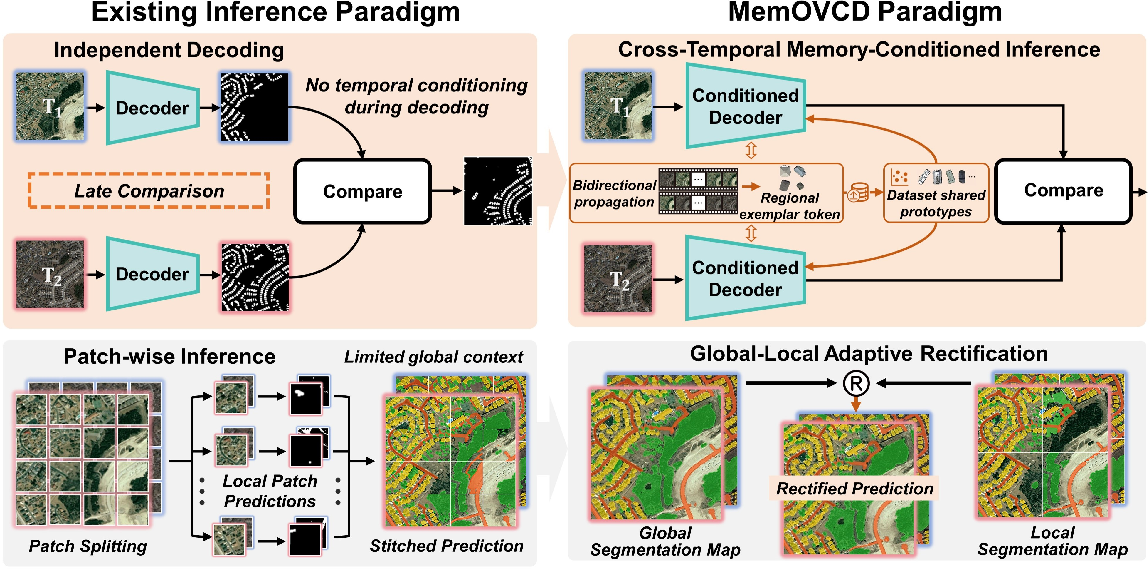}
\caption{Motivation of MemOVCD. Propagated regional features and shared prototypes guide semantic decoding before the two timestamps are compared. For large images, rectification fuses whole-image and stitched patch-wise predictions.}\label{fig1}
\end{figure}

Advances in vision foundation models have facilitated OVCD by introducing transferable visual representations and open-vocabulary semantic alignment, reducing the dependence on category-specific model training. DynamicEarth introduces text-driven change understanding by integrating SAM, DINO and CLIP into a training-free pipeline \citep{li2026dynamicearth}, while OmniOVCD streamlines the framework through a unified SAM 3-centered design \citep{zhang2026omniovcd}. In these representative pipelines, query-specific semantic predictions are obtained separately for each timestamp and subsequently compared. Temporal interaction at this stage does not condition the preceding semantic decoding.

However, this decoupling can limit the use of cross-temporal evidence in OVCD, as illustrated in Fig.~\ref{fig1}. Without cross-temporal semantic interaction at inference, appearance variations caused by seasonal changes, illumination differences or viewpoint shifts may be misinterpreted as semantic changes. Although the final comparison stage can identify inconsistencies between two independently generated predictions, it occurs after semantic interpretation and does not revise the information used to decode either timestamp. Consequently, semantic errors may persist and degrade the final change predictions. Additionally, high-resolution remote sensing images are commonly processed in a patch-wise manner, which often weakens global semantic continuity and degrades spatial consistency around patch boundaries, especially for large or irregular objects.

To overcome these challenges, we put forward MemOVCD, a training-free framework that explicitly introduces cross-temporal semantic interaction into the inference process. Specifically, we develop cross-temporal memory-conditioned inference (CTMC), which exploits the native memory mechanism of video SAM to propagate bidirectional semantic evidence between timestamps. Propagated candidate regions are first pooled over target-image features, and the resulting region features are then averaged into local visual exemplar tokens. Since local candidate regions can be absent, we further aggregate regional features across the unlabeled target collection into shared prototypes, whose number increases with the proportion of pairs lacking local exemplars. Local exemplars and shared prototypes will jointly guide subsequent semantic segmentation. To bridge appearance differences during propagation, we construct a synthetic transition-frame strategy. To improve prediction quality in high-resolution scenes, we introduce a global-local rectification strategy that adaptively fuses local and global predictions according to connected-component characteristics. Our main contributions are as follows.

\begin{figure}[!t]\centering
\includegraphics[width=\linewidth]{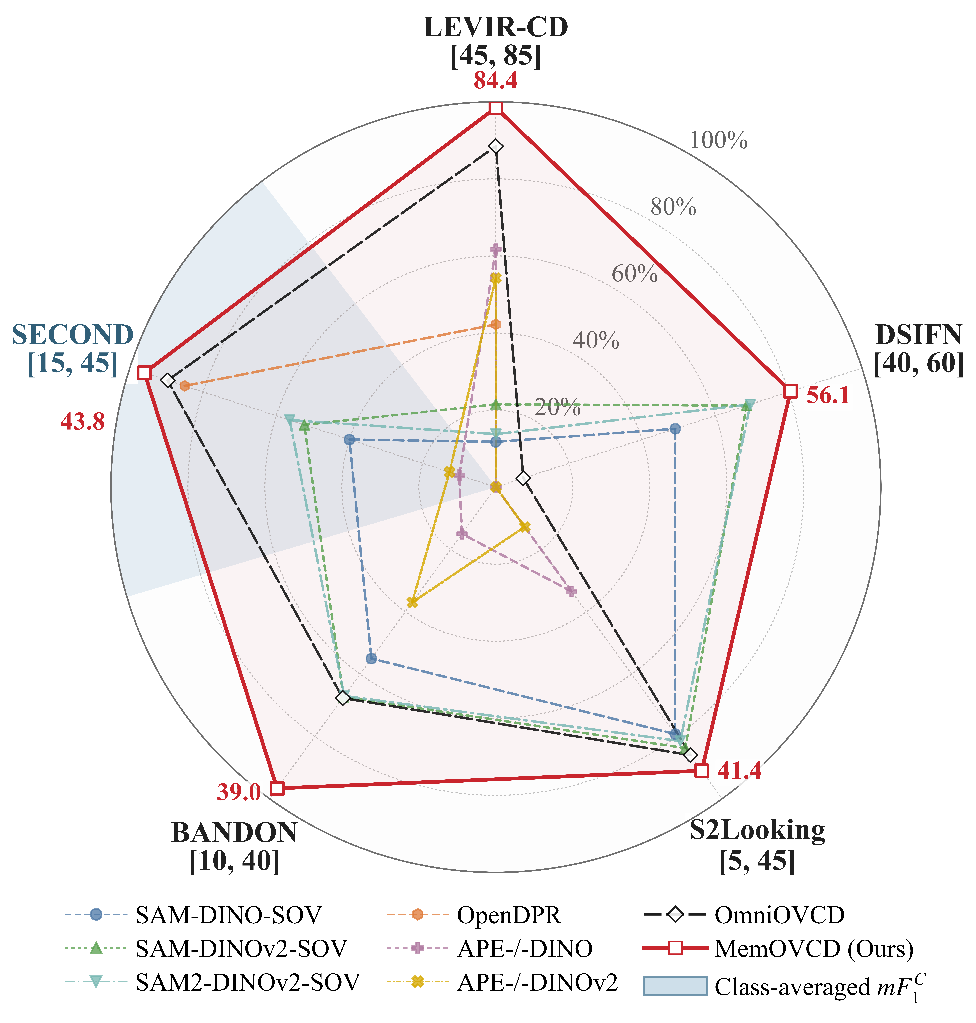}
\caption{$F^c_1$ across five benchmarks. Brackets give the lower and upper bounds of each normalized axis.}\label{fig2}
\end{figure}

\begin{itemize}

\item

We revisit open-vocabulary change detection from the perspective of explicit cross-temporal semantic interaction and formulate a memory-conditioned inference framework that departs from conventional independent bitemporal interpretation.

\item

We introduce CTMC to jointly condition semantic decoding on timestamp-specific visual exemplars and cross-image shared prototypes, with the shared prototype count determined by exemplar availability.

\item

We introduce a global-local adaptive rectification strategy to mitigate spatially incomplete predictions caused by patch-wise inference.

\item

Extensive experiments on five benchmarks demonstrate the effectiveness of MemOVCD, with Fig.~\ref{fig2} exhibiting its competitive overall performance.

\end{itemize}

\section{Related Work}

\subsection{Foundation Models for Open-Vocabulary Remote Sensing}

The rapid development of vision foundation models has substantially advanced open-vocabulary remote sensing by providing transferable semantic priors without task-specific retraining. Vision-language models represented by CLIP \citep{CLIP} enable semantic recognition beyond predefined label spaces through image-text alignment, while self-supervised visual encoders such as DINO \citep{DINOv3} provide discriminative visual representations for dense prediction. Meanwhile, the Segment Anything Model (SAM) family has established a promptable segmentation paradigm that decouples object localization from semantic recognition. SAM \citep{SAMv1} and SAM 2 \citep{SAMv2} support flexible interactive segmentation from various prompt types, whereas SAM 3 \citep{SAMv3} further extends this capability with unified image and video modeling. These models support remote sensing applications including open-vocabulary detection, segmentation and visual grounding \citep{guo2026dual,hwang2025fase,li2026exploring,li2026rsvg,li2025segearth,pan2025locate,zhan2025skyeyegpt}. Nevertheless, directly transferring foundation models to remote sensing remains challenging due to large appearance variations, high-resolution imagery and domain-specific semantic characteristics \citep{chen2025dynamicvis,huo2025remote,liu2024remoteclip}.
These challenges have also motivated domain-specific representation learning
\citep{kuang2025ecp,kuang2023complex,bi2019polsar,bi2017polsar,mao2024thz,liu2025hsi}.

\subsection{Open-Vocabulary Change Detection}

OVCD extends conventional change detection by identifying semantic changes specified through language queries. Existing works can be divided into training-required and training-free approaches. Training-required methods improve change understanding through additional optimization on remote sensing datasets \citep{guo2026opendpr,zhu2025univcd,zhuang2025ov}. Training-free OVCD has emerged as a practical alternative via composing pretrained foundation models. Representative methods explore different system designs, including collaborative foundation-model pipelines, adaptive information fusion, proposal-enhanced localization, unified SAM 3-based architectures and segmentation model-based reformulation \citep{dou2026adaptovcd,giannikos2026knowing,guo2026opendpr,li2026dynamicearth,su2026seg2change,zhang2026omniovcd}. CoRegOVCD calibrates semantic predictions and compares them using geometric and regional constraints \citep{tang2026coregovcd}. ReA-OVCD and OVRC perform post-decoding reliability assessment and semantic-guided recall calibration \citep{cao2026ovrc,zhu2026rea}. While these methods use temporal information to compare or refine predictions, our distinction mainly concerns how propagated visual evidence guides semantic decoding rather than temporal interaction alone. Fully supervised work \citep{zhu2025change3d} has laid a foundation for video-based change detection with learnable perception frames. MemOVCD instead targets training-free OVCD in remote sensing, combining query-specific visual exemplar features from propagated regions with shared prototypes to guide semantic inference.

\section{Methods}

\begin{figure*}[!t]\centering
\includegraphics[width=\linewidth]{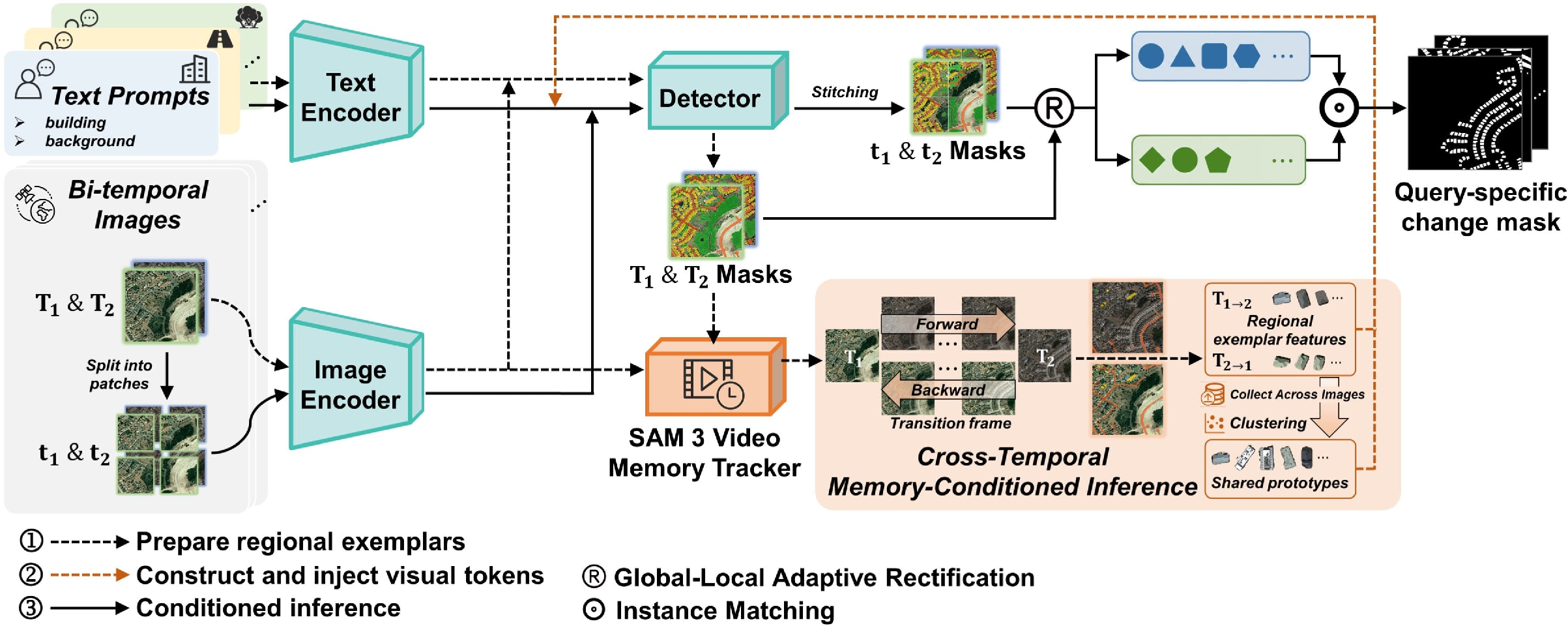}
\caption{Overview of MemOVCD. Features pooled from propagated regions form local exemplars and shared prototypes that jointly guide segmentation at both timestamps. Small images use whole-image decoding. Large images use patch-wise decoding and global-local adaptive rectification.}\label{fig3}
\end{figure*}

\subsection{Cross-Temporal Memory-Conditioned Inference}

Given co-registered images $T_1,T_2\in\mathbb R^{H\times W\times3}$ and a text query $q$, training-free OVCD predicts a binary mask $Y^q$ indicating changed locations that match $q$. Pretrained foundation models implement the mapping $(T_1,T_2,q)\mapsto Y^q$. Most methods decode semantics independently with $h$ and apply a late comparison $c$:
\begin{equation}
Y^q=c\big(h(T_1,q),h(T_2,q)\big).
\end{equation}
We instead condition semantic decoding on cross-temporal evidence:
\begin{equation}
Y^q=c\big(\phi(T_1\mid T_2,q,\mathcal B_q),\phi(T_2\mid T_1,q,\mathcal B_q)\big).
\end{equation}
The shared visual memory $\mathcal B_q$ is constructed from an unlabeled collection $\mathcal D_q$ of $N$ image pairs. We use the complete target dataset, fix the memory before final inference, and keep all network weights frozen.

\textbf{Bidirectional propagation.}
To bridge appearance gaps, we construct a short histogram-aligned sequence
\begin{equation}
V_K(T_1,T_2)=\{T_1,\widehat T^{(1)},\ldots,\widehat T^{(K)},T_2\},
\end{equation}
with intermediate frames
\begin{equation}
\widehat T^{(k)}=(1-\lambda_k)\mathcal H(T_1\mid T_2)+\lambda_k T_2,\quad
\lambda_k=\frac{k}{K+1}.
\end{equation}
Channel-wise histogram matching $\mathcal H$ adjusts global appearance statistics while preserving spatial structure. We build the reverse sequence separately, keeping the original images as endpoints. Query-specific masks $M_1^q,M_2^q$ and query $q$ initialize a frozen video tracker $\Phi$:
\begin{equation}
\begin{aligned}
S_2^q&=\Phi(q,M_1^q,V_K(T_1,T_2)),\\
S_1^q&=\Phi(q,M_2^q,V_K(T_2,T_1)).
\end{aligned}
\end{equation}
We retain nonempty target-endpoint masks. SAM 3 filters suppressed, removed and unconfirmed objects. When the source mask is empty, the text query alone initializes the tracker.

\textbf{Local visual exemplar.}
Let $F_t$ denote the full-image backbone feature map and $S^q_{t,r}$ a candidate region. Map each original foreground pixel to the feature grid using $\pi(i,j)=(\lfloor iH_f/H\rfloor,\lfloor jW_f/W\rfloor)$. For $R_t$ valid regional vectors, compute
\begin{equation}
z^q_{t,r}=\frac{1}{|S^q_{t,r}|}\sum_{(i,j)\in S^q_{t,r}}F_t[:,\pi(i,j)],\quad
e_t^q=\frac{1}{R_t}\sum_{r=1}^{R_t}z^q_{t,r}.
\end{equation}
Each region contributes equally to the timestamp-specific exemplar.

\textbf{Shared regional memory.}
A timestamp may lack a local exemplar ($R_t=0$). We pool the original regional vectors across $\mathcal D_q$. Let $A$ count pairs with at least one valid region at either timestamp. The empirical prototype count is
\begin{equation}
p=A/N,\qquad K_s=\max\{1,\lceil5(1-p)\rceil\}.
\end{equation}
We set $p=0$ when $A=0$. The rule allocates more prototypes when local evidence is scarce. Weighted Euclidean $k$-means++ produces prototypes $\{c_k^q\}_{k=1}^{K_s}$. Each region has weight proportional to $1/(T_i R_{i,t})$, where $T_i$ counts nonempty timestamps for pair $i$ and $R_{i,t}$ counts retained regions at timestamp $t$. Each contributing pair thus receives equal total weight.

For each foreground prompt we concatenate text tokens, the optional local exemplar, and the shared tokens: $P_t^q=[T^q;e_t^q\ (\text{if available});c_1^q;\ldots;c_{K_s}^q]$. Image features attend to these tokens in SAM 3's existing cross-attention layers before mask decoding. Background prompts receive no visual tokens.

\subsection{Global-Local Adaptive Rectification}
Patch-wise inference preserves small targets but can fragment predictions at patch boundaries. For timestamp $t$, let $s_t^l$ denote the stitched patch foreground scores conditioned on $P_t^q$, and $s_t^g$ the initial text-only whole-image scores. Their binary masks are $B_t^l$ and $B_t^g$. For each 8-connected component $C_{t,r}$ of $B_t^g$ with area at least $A_{\min}$, define $\rho_{t,r}=|C_{t,r}\cap B_t^l|/|C_{t,r}|$. The global weight is
\begin{equation}
\alpha(\rho)=\begin{cases}
1,&\rho<\tau_{\rm miss},\\
\frac{\tau_{\rm keep}-\rho}{\tau_{\rm keep}-\tau_{\rm miss}},&\tau_{\rm miss}\le\rho<\tau_{\rm keep},\\
0,&\rho\ge\tau_{\rm keep}.
\end{cases}
\end{equation}
Here $\tau_{\rm miss}<\tau_{\rm keep}$ are local-coverage thresholds. Initialize $s_t=s_t^l$. For $x\in C_{t,r}$, update $s_t(x)=\alpha(\rho_{t,r})s_t^g(x)+[1-\alpha(\rho_{t,r})]s_t^l(x)$. Reapply foreground--background competition and threshold $\tau$. The global prediction contributes more when local coverage is insufficient.

\subsection{Inference Pipeline}
\textbf{Preparation and memory construction.} Initial text-only whole-image segmentation, bidirectional propagation and coordinate-based feature pooling run across the unlabeled target collection. Regional features construct the shared memory, which remains fixed throughout final inference.

\textbf{Visually guided decoding.} Images fitting in a single $512\times512$ patch are re-encoded and decoded with $P_t^q$ without rectification. Larger images are padded and split into nonoverlapping $512\times512$ patches, each decoded with $P_t^q$. After stitching and cropping, rectification fuses the local scores with the original text-only global scores.

\textbf{Semantic prediction.} Foreground scores combine semantic and instance predictions with presence confidence. Among synonymous foreground prompts, we select the prediction with the highest presence score. We maximize background scores pixelwise across background prompts. A foreground pixel must exceed both the background score and $\tau$.

\textbf{Change prediction.} For each 8-connected foreground region $R$, let $M$ denote the entire opposite-timestamp mask. We mark the whole region as changed if $o(R,M)=|R\cap M|/|R|<m$. The union of regions marked in either direction gives $Y^q$.

\section{Experiments}

\subsection{Experimental Settings}

\textbf{Datasets.} We evaluate our method on five benchmarks, including LEVIR-CD \citep{chen2020spatial}, DSIFN \citep{zhang2020deeply}, S2Looking \citep{shen2021s2looking} and BANDON \citep{pang2023detecting} for binary change detection, and SECOND \citep{yang2021asymmetric} for semantic change detection.

\textbf{Evaluation metrics.} We report the F1-score ($F^c_1$) and intersection over union ($\mathrm{IoU}^c$) of true changes. For semantic change detection, we also report their class means ($\mathrm{mF}^c_1$ and $\mathrm{mIoU}^c$). All results are percentages unless otherwise specified.

\textbf{Implementation Details.} We employ SAM 3.1 as our backbone with frozen weights on an A100 GPU. Thresholds $\tau$ and $m$ are selected for each dataset or semantic category. As text prompt design is not the focus of MemOVCD, we follow the synonym-based prompting strategy of \citet{li2025segearth,zhang2026omniovcd}, and keep prompt lists fixed for fair comparisons.

\subsection{Main Results}

\textbf{Compared Methods.} Compared methods are grouped into traditional unsupervised, training-required and training-free approaches. OVCD methods can be further divided into Mask-Compare-Identify (M-C-I), Identify-Mask-Compare (I-M-C), and other frameworks. The traditional unsupervised methods include IRMAD \citep{nielsen2007regularized}, PCA-Kmeans \citep{celik2009unsupervised}, ISFA \citep{wu2013slow}, DCVA \citep{saha2019unsupervised} and DSFA \citep{DSFA}. Among training-free methods, the M-C-I group includes AnyChange \citep{zheng2024segment}, AdaptOVCD \citep{dou2026adaptovcd}, OpenDPR \citep{guo2026opendpr}, and three DynamicEarth variants, namely SAM-DINO-SOV, SAM-DINOv2-SOV and SAM2-DINOv2-SOV, where SOV is the abbreviation of SegEarth-OV. The I-M-C group includes OmniOVCD \citep{zhang2026omniovcd} and DynamicEarth variants APE-/-DINO and APE-/-DINOv2. For reference, we also include training-based methods such as UniVCD \citep{zhu2025univcd}, OpenDPR-W \citep{guo2026opendpr} and Seg2Change \citep{su2026seg2change}. Since MemOVCD uses shared prototypes under a transductive protocol, we additionally present an inductive variant, L-MemOVCD, which relies solely on local exemplars, to ensure a fair comparison. Rankings include only inductive training-free OVCD methods.

\begin{table}[!t]\centering
\caption{Comparison on building change detection. Asterisks denote our reimplementations under the best settings. Dashes indicate unavailable values. L-MemOVCD uses only local exemplars. Rankings cover inductive training-free methods. Bold underline, bold and underline denote first, second and third.}\label{tab1}
\resizebox{\linewidth}{!}{\begin{tabular}{lcccccccc}\toprule
Method & \multicolumn{2}{c}{LEVIR-CD} & \multicolumn{2}{c}{DSIFN} & \multicolumn{2}{c}{S2Looking} & \multicolumn{2}{c}{BANDON} \\
 & $\mathrm{IoU}^c$ & $F_1^c$ & $\mathrm{IoU}^c$ & $F_1^c$ & $\mathrm{IoU}^c$ & $F_1^c$ & $\mathrm{IoU}^c$ & $F_1^c$\\\midrule
\multicolumn{9}{c}{\emph{Traditional unsupervised methods}} \\
IRMAD & 6.7 & 12.6 & 6.0 & 11.3 & - & - & - & - \\
PCA-Kmeans & 5.4 & 10.2 & 20.1 & 33.4 & - & - & - & - \\
ISFA & 5.4 & 10.2 & 17.4 & 29.7 & - & - & - & - \\
DCVA & 13.3 & 24.5 & 17.9 & 30.4 & - & - & - & - \\
DSFA & 7.9 & 14.7 & 18.2 & 30.8 & - & - & - & - \\
\multicolumn{9}{c}{\emph{Training-required methods}} \\
\multicolumn{9}{c}{\emph{M-C-I}} \\
UniVCD & 56.7 & 72.3 & - & - & - & - & 17.7 & 30.1 \\
OpenDPR-W & 48.1 & 65.0 & - & - & - & - & - & - \\
\multicolumn{9}{c}{\emph{Other Frameworks}} \\
Seg2Change & 64.9 & 78.7 & 41.4 & 58.6 & - & - & - & - \\
\multicolumn{9}{c}{\emph{Training-free OVCD methods}} \\
\multicolumn{9}{c}{\emph{M-C-I}} \\
SAM-DINO-SOV & 33.0 & 49.7 & 33.1* & 49.8* & 22.5 & 36.7 & 15.3 & 26.5 \\
SAM-DINOv2-SOV & 36.6 & 53.6 & 36.7* & 53.7* & \underline{23.9} & \underline{38.5} & 17.6 & \underline{30.2} \\
SAM2-DINOv2-SOV & 33.8 & 50.5 & \underline{36.9*} & \underline{53.9*} & 23.1 & 37.6 & \underline{17.7} & 30.1 \\
AnyChange & 19.5 & 32.7 & 24.4 & 39.2 & - & - & - & - \\
AdaptOVCD & 51.5 & 68.0 & \underline{\textbf{42.3}} & \underline{\textbf{59.5}} & - & - & - & - \\
OpenDPR & 44.8 & 61.9 & - & - & - & - & - & - \\
\multicolumn{9}{c}{\emph{I-M-C}} \\
APE - / - DINO & \underline{53.5} & \underline{69.7} & 12.1* & 21.6* & 10.1 & 18.4 & 7.8 & 14.5 \\
APE - / - DINOv2 & 50.0 & 66.7 & 15.2* & 26.4* & 5.3 & 10.1 & 11.8 & 21.1 \\
OmniOVCD & \textbf{67.2} & \textbf{80.4} & 26.2* & 41.5* & \textbf{24.5} & \textbf{39.4} & \textbf{17.8*} & \textbf{30.3*} \\
L-MemOVCD & \underline{\textbf{71.9}} & \underline{\textbf{83.6}} & \textbf{37.4} & \textbf{54.5} & \underline{\textbf{25.6}} & \underline{\textbf{40.7}} & \underline{\textbf{23.7}} & \underline{\textbf{38.3}} \\
MemOVCD & 73.0 & 84.4 & 39.0 & 56.1 & 26.1 & 41.4 & 24.2 & 39.0 \\\bottomrule\end{tabular}}
\end{table}

\begin{table*}[!t]\centering
\caption{Comparison on SECOND. $\dagger$ denotes results from the original paper. $\approx0$ denotes a near-zero value. L-MemOVCD uses only local exemplars. Rankings cover inductive training-free methods. Bold underline, bold and underline denote first, second and third.}\label{tab2}
\resizebox{\linewidth}{!}{\begin{tabular}{lcccccccccccccc}\toprule
Method & \multicolumn{2}{c}{Building} & \multicolumn{2}{c}{Tree} & \multicolumn{2}{c}{Water} & \multicolumn{2}{c}{Low veg.} & \multicolumn{2}{c}{Surface} & \multicolumn{2}{c}{Playground} & \multicolumn{2}{c}{Class avg.} \\
 & $\mathrm{IoU}^c$ & $F_1^c$ & $\mathrm{IoU}^c$ & $F_1^c$ & $\mathrm{IoU}^c$ & $F_1^c$ & $\mathrm{IoU}^c$ & $F_1^c$ & $\mathrm{IoU}^c$ & $F_1^c$ & $\mathrm{IoU}^c$ & $F_1^c$ & $\mathrm{mIoU}^c$ & $\mathrm{mF}^c_1$\\\midrule
SAM-DINO-SOV & 35.2 & 52.1 & 8.7 & 16.1 & 1.9 & 3.7 & 20.0 & 33.3 & 21.8 & 35.8 & 11.8 & 21.0 & 16.6 & 27.0 \\
SAM-DINOv2-SOV & 38.8 & 55.9 & 9.8 & 17.8 & 4.5 & 8.5 & 21.8 & 35.8 & 25.1 & 40.1 & 15.0 & 26.1 & 19.1 & 30.7 \\
SAM2-DINOv2-SOV & 36.7 & 53.7 & 10.7 & 19.3 & 1.8 & 3.6 & 20.8 & 34.5 & 19.4 & 32.5 & \textbf{31.6} & \textbf{48.0} & 20.2 & 31.9 \\
APE - / - DINO & 27.4 & 43.0 & 9.7 & 17.7 & \underline{12.3} & \underline{22.0} & 0.1 & 0.1 & $\approx0$ & $\approx0$ & 14.3 & 25.0 & 10.6 & 18.0 \\
APE - / - DINOv2 & 29.9 & 46.0 & 10.3 & 18.7 & \underline{12.3} & \underline{22.0} & 0.4 & 0.8 & $\approx0$ & $\approx0$ & 14.6 & 25.4 & 11.3 & 18.8 \\
OpenDPR$^{\dagger}$ & \underline{43.2} & \underline{60.4} & \underline{\textbf{19.7}} & \underline{\textbf{33.0}} & 12.0 & 21.4 & \underline{23.9} & \underline{38.6} & \underline{\textbf{31.2}} & \underline{\textbf{47.5}} & 26.5 & 42.0 & \underline{26.1} & \underline{40.5} \\
OmniOVCD & \textbf{45.2} & \textbf{62.3} & \underline{16.6} & \underline{28.5} & \textbf{21.2} & \textbf{35.0} & \textbf{24.5} & \textbf{39.3} & \textbf{27.7} & \textbf{43.4} & \underline{27.2} & \underline{42.7} & \textbf{27.1} & \textbf{41.9} \\
L-MemOVCD & \underline{\textbf{47.6}} & \underline{\textbf{64.5}} & \textbf{17.2} & \textbf{29.3} & \underline{\textbf{21.8}} & \underline{\textbf{35.8}} & \underline{\textbf{25.0}} & \underline{\textbf{40.0}} & \underline{27.4} & \underline{43.0} & \underline{\textbf{33.1}} & \underline{\textbf{49.7}} & \underline{\textbf{28.7}} & \underline{\textbf{43.7}} \\
MemOVCD & 47.9 & 64.8 & 18.0 & 30.5 & 23.9 & 38.6 & 24.5 & 39.3 & 28.2 & 44.0 & 29.7 & 45.8 & 28.7 & 43.8 \\\bottomrule\end{tabular}}
\end{table*}

\textbf{Binary Change Detection.} As reported in Tab.~\ref{tab1}, L-MemOVCD ranks first on LEVIR-CD, S2Looking and BANDON, and second on DSIFN among all compared training-free OVCD methods. On LEVIR-CD, it exceeds OmniOVCD by 4.7 points in $\mathrm{IoU}^{c}$ and 3.2 points in $F^{c}_{1}$. For reference, it also exceeds Seg2Change, the strongest training-required reported competitor on LEVIR-CD. The corresponding gains are 1.1 and 1.3 points on S2Looking, and 5.9 and 8.0 points on BANDON. Although AdaptOVCD achieves the highest scores on DSIFN, L-MemOVCD outperforms all listed training-free I-M-C competitors. Overall, the results support the effectiveness of MemOVCD as a training-free framework for building change detection.

\textbf{Land Cover Change Detection.} As reported in Tab.~\ref{tab2}, L-MemOVCD achieves the best class averages among the compared methods on SECOND, with 28.7\% $\mathrm{mIoU}^{c}$ and 43.7\% $\mathrm{mF}^{c}_{1}$, outperforming OmniOVCD by 1.6 and 1.8 points respectively. It ranks first on building, water, low vegetation and playground, second on tree, and third on surface. On playground, it exceeds SAM2-DINOv2-SOV by 1.5 points in $\mathrm{IoU}^{c}$ and 1.7 points in $F^{c}_{1}$. Although OpenDPR, which employs additional diffusion-generated visual references and its original prompt settings, leads on tree and surface, our method remains competitive across all six categories. With shared memory, MemOVCD obtains 28.7\% $\mathrm{mIoU}^{c}$ and 43.8\% $\mathrm{mF}^{c}_{1}$ under the transductive protocol. These results support the applicability of MemOVCD to land cover changes.

\subsection{Ablation Study}

\begin{table*}[!t]\centering
\caption{Component ablation. Local, Shared and Adapt. indicate local exemplars, shared prototypes and adaptive prototype counts. S-CTMC uses one shared prototype. Rectification is omitted on DSIFN and SECOND. $\ddagger$ marks L-MemOVCD results.}\label{tab3}
\resizebox{\linewidth}{!}{\begin{tabular}{lcccccccccccccc}\toprule
Model & Local & Shared & Adapt. & Recti. & \multicolumn{2}{c}{LEVIR-CD} & \multicolumn{2}{c}{DSIFN} & \multicolumn{2}{c}{S2Looking} & \multicolumn{2}{c}{BANDON} & \multicolumn{2}{c}{SECOND} \\ & & & & & $\mathrm{IoU}^c$ & $F^c_1$ & $\mathrm{IoU}^c$ & $F^c_1$ & $\mathrm{IoU}^c$ & $F^c_1$ & $\mathrm{IoU}^c$ & $F^c_1$ & $\mathrm{mIoU}^c$ & $\mathrm{mF}^c_1$\\\midrule
Base & \xmark & \xmark & \xmark & \xmark & 70.14 & 82.45 & 37.85 & 54.92 & 22.90 & 37.26 & 22.49 & 36.72 & 28.50 & 43.51 \\
L-CTMC & \cmark & \xmark & \xmark & \xmark & 70.52 & 82.71 & 37.41 $^{\ddagger}$ & 54.45$^{\ddagger}$ & 23.52 & 38.08 & 22.82 & 37.16 & 28.67 $^{\ddagger}$ & 43.72$^{\ddagger}$ \\
S-CTMC & \cmark & \cmark & \xmark & \xmark & 71.70 & 83.52 & 37.90 & 54.96 & 24.17 & 38.93 & 23.30 & 37.79 & 28.78 & 43.87 \\
CTMC & \cmark & \cmark & \cmark & \xmark & 71.70 & 83.52 & 38.97 & 56.08 & 24.17 & 38.93 & 23.30 & 37.79 & 28.71 & 43.85 \\
Base + Recti & \xmark & \xmark & \xmark & \cmark & 71.58 & 83.44 & - & - & 25.09 & 40.11 & 23.55 & 38.12 & - & - \\
L-CTMC + Recti & \cmark & \xmark & \xmark & \cmark & 71.89 $^{\ddagger}$ & 83.64$^{\ddagger}$ & - & - & 25.58 $^{\ddagger}$ & 40.73$^{\ddagger}$ & 23.68$^{\ddagger}$ & 38.29$^{\ddagger}$ & - & - \\
MemOVCD & \cmark & \cmark & \cmark & \cmark & 72.97 & 84.37 & - & - & 26.13 & 41.43 & 24.25 & 39.03 & - & - \\\bottomrule\end{tabular}}
\end{table*}

As shown in Tab.~\ref{tab3}, we conduct ablation studies to evaluate each component of MemOVCD. Starting from a plain training-free I-M-C Base, we progressively introduce local exemplars and shared memory through CTMC, followed by global-local adaptive rectification.

Local exemplars improve performance on four of the five benchmarks, with only a marginal drop on DSIFN. L-CTMC relies solely on inductive intra-pair local memory, whereas S-CTMC adopts a transductive protocol by building shared prototypes from the unlabeled target collection. Under identical conditions, S-CTMC yields consistent improvements across all five datasets, demonstrating that these gains stem from unlabeled visual information.

Furthermore, compared with a single shared prototype, the adaptive rule in CTMC brings a notable gain in $\mathrm{IoU}^{c}$ on DSIFN. However, it slightly underperforms S-CTMC on SECOND, indicating that increasing the prototype count does not guarantee better performance. On LEVIR-CD, S2Looking and BANDON, S-CTMC and CTMC yield identical results, as adaptive rule selects only 1 prototype.

Recti. further boosts CTMC on large-image datasets, improving $\mathrm{IoU}^{c}$ by 1.27, 1.93 and 0.98 points on LEVIR-CD, S2Looking and BANDON, respectively. Note that MemOVCD also consistently outperforms both Base + Recti. and L-CTMC + Recti., validating the complementary contributions of shared memory and spatial rectification.

\subsection{Computation Cost and Efficiency Analysis}

\begin{figure}[!t]\centering
\includegraphics[width=\linewidth]{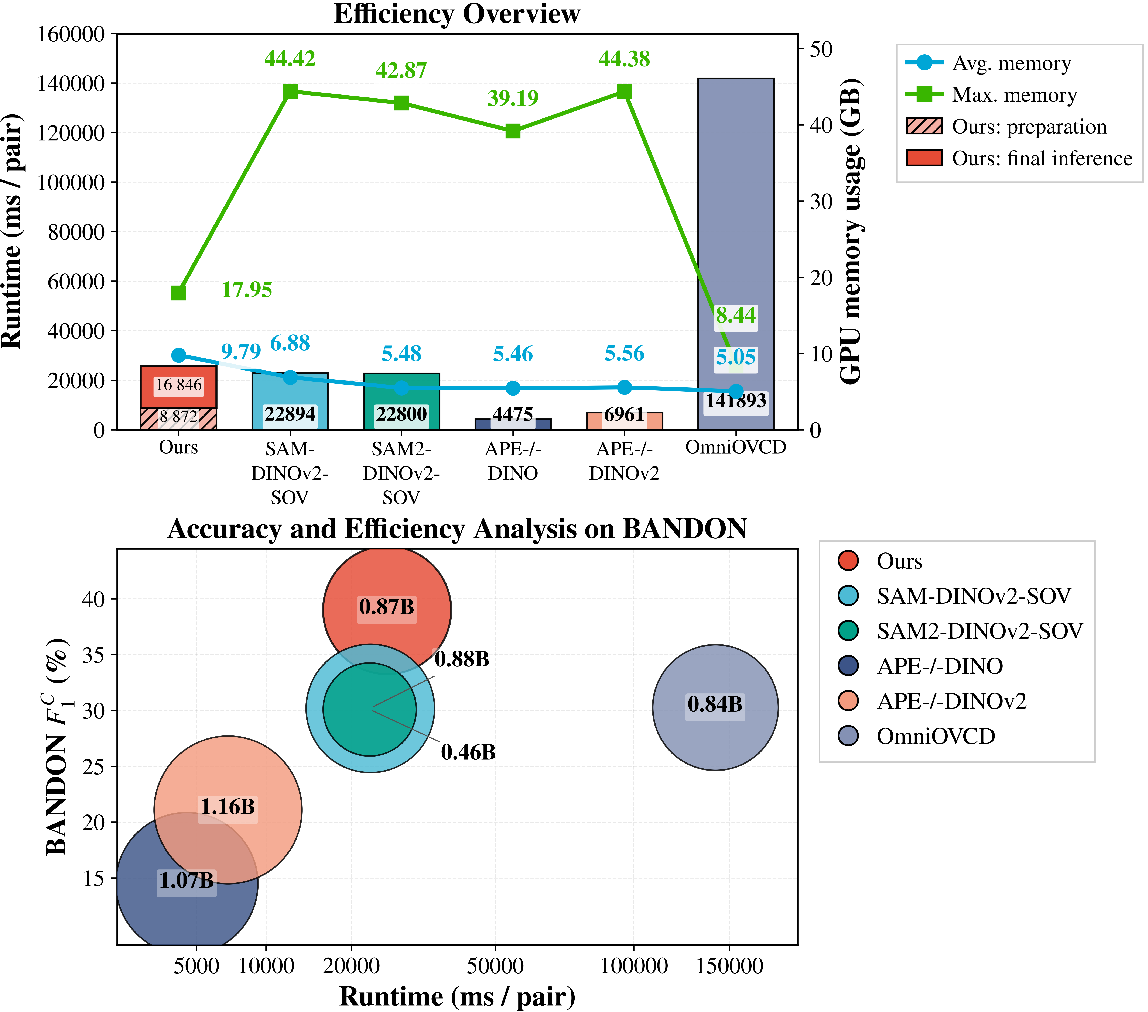}
\caption{Efficiency comparison on BANDON. Top: runtime and GPU memory per image pair. MemOVCD runtime includes preparation and final inference. Mean memory is a time-weighted estimate. Bottom: accuracy--efficiency trade-off. Bubble size denotes parameter count.}\label{fig4}
\end{figure}

We reproduce six methods using their official codes and evaluate their computational cost and efficiency on BANDON on the same A100 GPU. As shown in Fig.~\ref{fig4}, DynamicEarth variants suffer from high peak memory because they upsample dense DINO-based features to the original high image resolution. OmniOVCD, on the other hand, performs pairwise comparisons between full-resolution instance masks, which leads to significant CPU overhead in scenes with dense instances.

By contrast, MemOVCD relies on patch-wise semantic inference and region-to-foreground comparison. This design effectively reduces peak memory compared to DynamicEarth and shortens runtime compared to OmniOVCD. Although the preparation stage and repeated encoding introduce some overhead, MemOVCD still maintains a reasonable accuracy-cost trade-off.

\subsection{Visualization}

\begin{figure}[!t]\centering
\includegraphics[width=\linewidth]{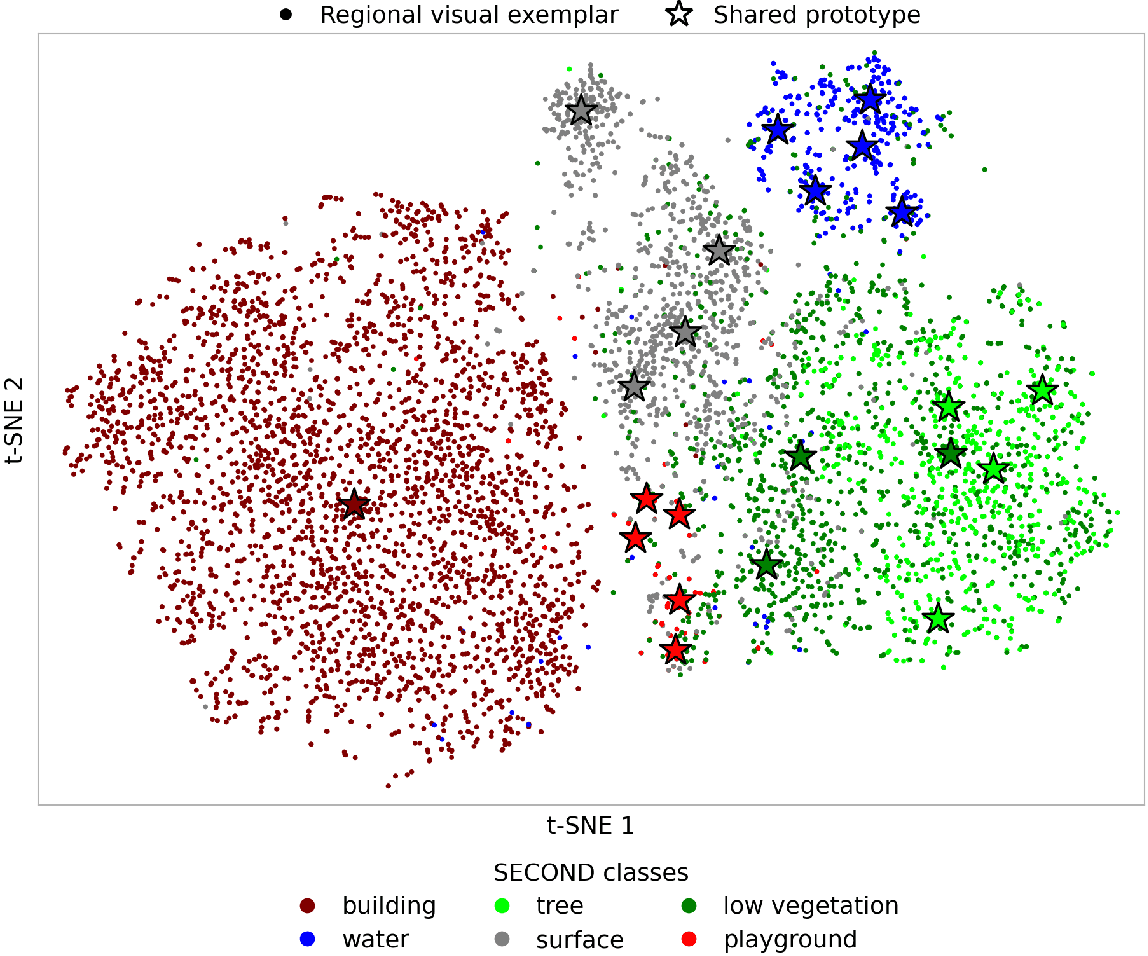}
\caption{Joint t-SNE visualization on SECOND. Colors follow the native SECOND color bar and indicate the query category.}\label{fig5}
\end{figure}

\textbf{Regional Exemplars and Shared Prototypes.} To examine how shared memory summarizes propagated regional features, we visualize the regional exemplars and shared prototypes on SECOND. All 7341 regional features and 22 shared prototypes from the full test set are jointly projected using PCA followed by t-SNE. As shown in Figure~\ref{fig5}, regional features exhibit a clear structural distribution across categories, with building and water forming relatively distinct groups, while tree and low vegetation overlap. The shared prototypes occupy different parts of these regional distributions, demonstrating their capability to aggregate features across different images and represent the inherent appearance variations within the target dataset. This visualization illustrates the distribution and representativeness of the shared memory.

\begin{figure}[!t]\centering
\includegraphics[width=\linewidth]{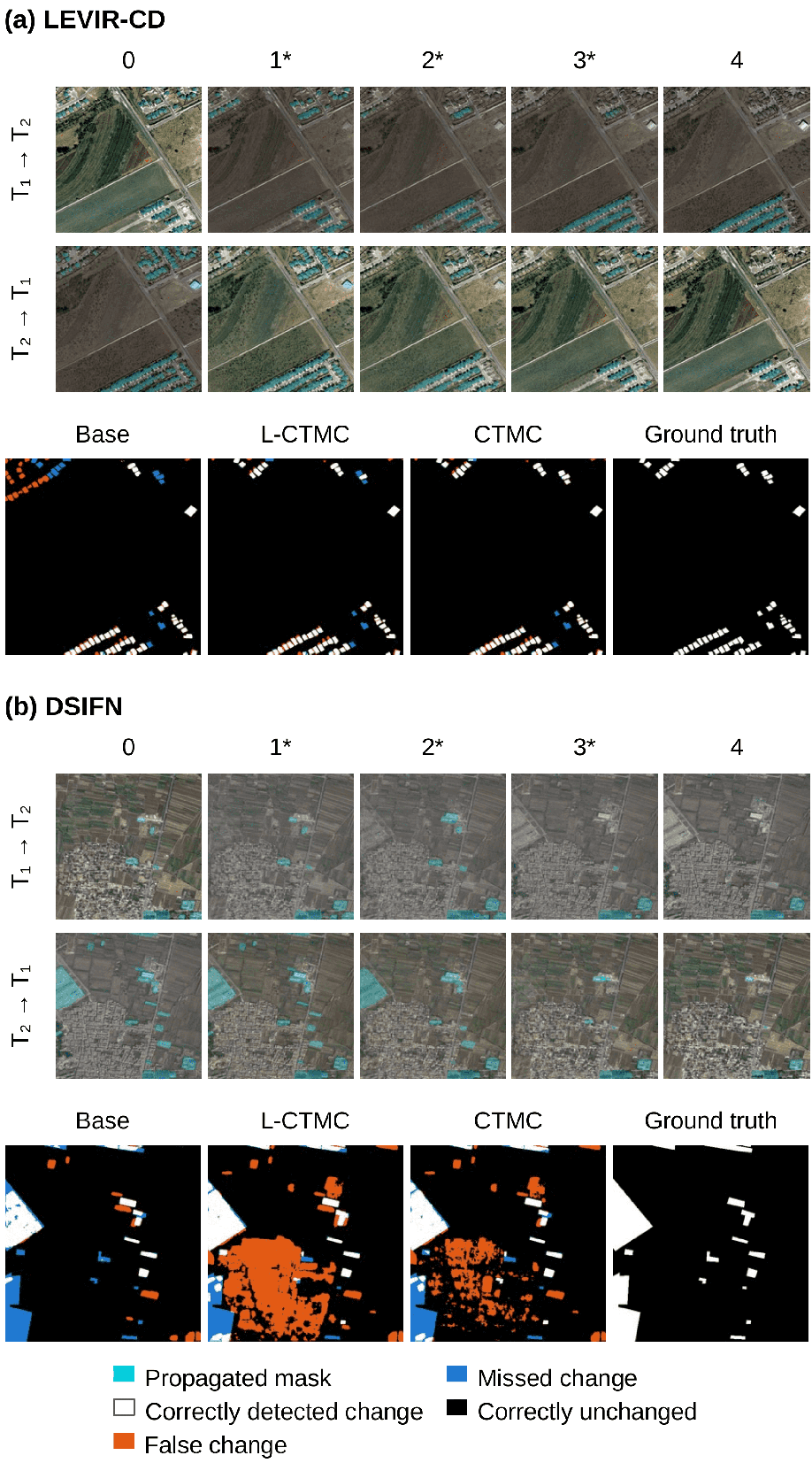}
\caption{Bidirectional propagation and change predictions on (a) LEVIR-CD and (b) DSIFN. Cyan overlays show propagated masks. Asterisks mark synthetic transitions.}\label{fig6}
\end{figure}

\textbf{Memory Propagation and Its Effect on Prediction.} Figure~\ref{fig6} illustrates masks from bidirectional propagation and the resulting change predictions, with Recti. disabled to isolate CTMC. We compare Base (no memory), L-CTMC (intra-pair local memory) and CTMC (cross-image shared memory). In LEVIR-CD case, CTMC successfully removes the false changes produced by Base while reducing the missed changes in L-CTMC, demonstrating the effect of cross-image shared memory. We also illustrate a failure case on DSIFN, where local memory introduces additional false changes that shared memory only partially removes.

\subsection{CTMC across OVCD Pipelines and Backbones}

\begin{table}[!t]\centering
\caption{Adaptation of CTMC to training-free OVCD pipelines on DSIFN. Enhanced settings incorporate CTMC. Weights identify the frozen backbone.}\label{tab4}
\resizebox{\linewidth}{!}{\begin{tabular}{lcccc}\toprule
Setting & Method & Weights & $\mathrm{IoU}^c$ & $F^c_1$\\\midrule
DynamicEarth I-M-C & Base & SAM 2.1 & 6.2 & 11.7 \\
DynamicEarth I-M-C & Enhanced & SAM 2.1 & 7.8 & 14.5 \\
DynamicEarth M-C-I & Base & SAM 2.1 & 36.9 & 53.9 \\
DynamicEarth M-C-I & Enhanced & SAM 2.1 & 42.4 & 59.6 \\
OmniOVCD & Base & SAM 3 & 26.2 & 41.5 \\
OmniOVCD & Enhanced & SAM 3 & 26.3 & 41.6 \\
MemOVCD & Enhanced & SAM 3 & 38.3 & 55.3 \\
MemOVCD & Enhanced & SAM 3.1 & 39.0 & 56.1 \\\bottomrule\end{tabular}}
\end{table}

We further examine CTMC across various training-free OVCD pipelines and segmentation backbones. For DynamicEarth \citep{li2026dynamicearth}, we adapt MM-Grounding-DINO-SAM2-DINO (I-M-C) and SAM2-DINOv2-SOV (M-C-I), retaining their text prompts and change-decision rules.

In I-M-C, local and shared memory guide positive-point selection. We allocate a point budget equal to the number of Base instances and distribute it according to detection confidence and memory similarity. Then selected points will replace box prompts for SAM 2 re-segmentation. In M-C-I, SAM 2 first generates label-free masks to initialize memory propagation. Regional features and shared prototypes then refine the mask-logit prompts within these propagated regions. Finally, SAM 2 re-decodes the original candidates prior to change comparison and semantic identification. For OmniOVCD \citep{zhang2026omniovcd}, we append local visual tokens to the foreground text features while retaining its native score fusion and change head. We also evaluate full MemOVCD with the original SAM 3 weights.

As shown in Tab.~\ref{tab4}, the adapted I-M-C and M-C-I pipelines improve $\mathrm{IoU}^{c}$ by 1.6 and 5.3 points respectively. Note that although OmniOVCD shows marginal gains since its original settings were already specifically tuned, we empirically find that further tuning the enhanced version can yield even higher improvements. Moreover, MemOVCD achieves comparable $\mathrm{IoU}^{c}$ using the original SAM 3 and SAM 3.1 respectively. Overall, these results demonstrate the promising potential of extending video propagation and memory-guided inference to broader OVCD frameworks.

\subsection{Representative Qualitative Results}

\begin{figure}[!t]\centering
\includegraphics[width=\linewidth]{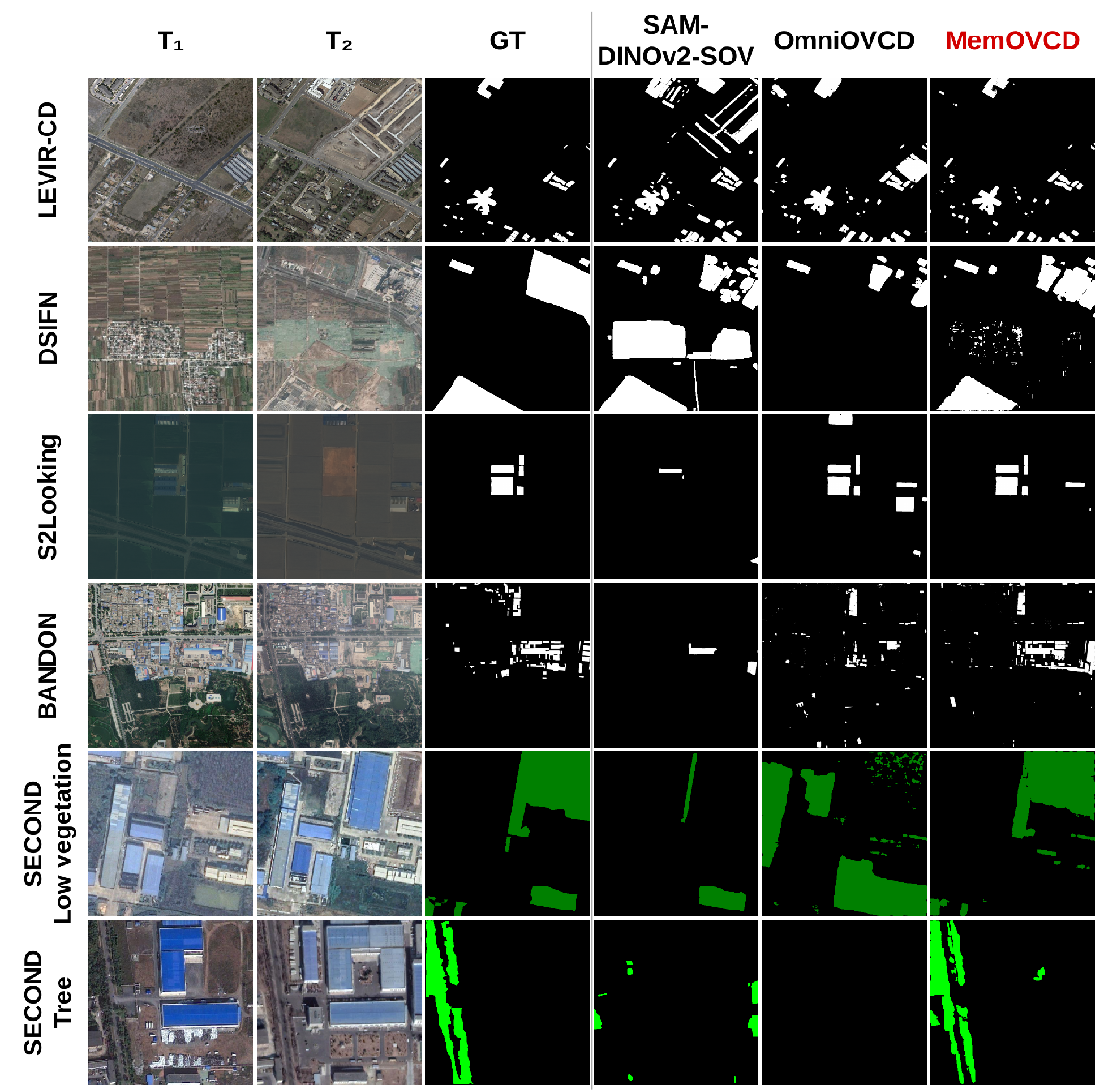}
\caption{Representative qualitative comparison of MemOVCD against two other methods on five datasets.}\label{fig7}
\end{figure}

Figure~\ref{fig7} compares representative qualitative results. SAM-DINOv2-SOV often misses compact changed targets, loses structural information within changed regions and produces fragmented predictions. OmniOVCD is more susceptible to false positives, suggesting insufficient calibration from local evidence, and exhibits inaccurate boundaries for categories with ambiguous shapes or contours. In comparison, our MemOVCD yields more complete and semantically coherent predictions overall while maintaining competitive robustness in challenging multi-class scenes. 

\FloatBarrier
\section{Conclusion}

This paper presented MemOVCD and L-MemOVCD, training-free frameworks which recast OVCD as cross-temporal memory-conditioned inference. Propagated candidate regions provide timestamp-specific local exemplars, while the shared prototype bank aggregates regional features to supply cross-image visual conditions. Both are jointly used to condition semantic segmentation. Furthermore, global-local adaptive rectification enhances spatial completeness for patch-wise inference. Experiments on five benchmarks indicate that stronger temporal coupling at inference stage is more effective than relying only on late comparison. These findings provide a practical direction for OVCD by incorporating cross-temporal conditioning and enhancing spatial completeness.

\bibliography{revised}

\end{document}